\documentclass[preprint,12pt]{elsarticle}

\usepackage{amssymb}
\usepackage{amsmath}

\usepackage[gen]{eurosym}
\usepackage{booktabs}  
\usepackage{makecell}  
\usepackage{graphicx}
\usepackage{subcaption}
\usepackage{layouts}
\usepackage{multirow}
\usepackage{cleveref}
\usepackage{siunitx}
\usepackage{amsmath}

\usepackage[acronym]{glossaries}
\newacronym{peft}{PEFT}{Parameter Efficient Fine-Tuning}
\newacronym{dff}{DFF}{Decision-Focused Fine-Tuning}
\newacronym{pff}{PFF}{Prediction-Focused Fine-Tuning}
\newacronym{mae}{MAE}{Mean Absolute Error}
\newacronym{mse}{MSE}{Mean Square Error}
\newacronym{dora}{DoRA}{Directional Rank Adaptation}
\newacronym{lora}{LoRA}{Low-Rank Adaption}
\newacronym{pv}{PV}{Photovoltaic}

\usepackage{siunitx}

\journal{Energy and AI}

\begin{document}

\begin{frontmatter}



\title{Decision-Focused Fine-Tuning of Time Series Foundation Models for Dispatchable Feeder Optimization} 

\renewcommand*{\today}{December 20, 2024}

\author[label1]{Maximilian Beichter} \ead{maximilian.beichter@kit.edu}
\author[label1,label2]{Nils Friederich}
\author[label1]{Janik Pinter}
\author[label1]{Dorina Werling}
\author[scc]{Kaleb Phipps}
\author[label1]{Sebastian Beichter}
\author[label1]{Oliver Neumann}
\author[label1]{Ralf Mikut}
\author[label1]{Veit Hagenmeyer}
\author[label3]{Benedikt Heidrich}

\affiliation[label1]{organization={Institute for Automation and Applied Informatics (IAI), Karlsruhe Institute of Technology (KIT)},
            addressline={Hermann-von-Helmholtz-Platz~1}, 
            city={Eggenstein-Leopoldshafen},
            postcode={76344}, 
            state={Baden-Württemberg},
            country={Germany}}

\affiliation[label2]{organization={Institute of Biological and Chemical Systems (IBCS), Karlsruhe Institute of Technology (KIT)},
            addressline={Hermann-von-Helmholtz-Platz~1}, 
            city={Eggenstein-Leopoldshafen},
            postcode={76344}, 
            state={Baden-Württemberg},
            country={Germany}}

\affiliation[scc]{organization={Scientific Computing Center (SCC), Karlsruhe Institute of Technology (KIT)},
            addressline={Hermann-von-Helmholtz-Platz~1}, 
            city={Eggenstein-Leopoldshafen},
            postcode={76344}, 
            state={Baden-Württemberg},
            country={Germany}}

\affiliation[label3]{organization={Mercedes-Benz~Tech~Innovation},
            addressline={Wilhelm-Runge-Straße~11}, 
            city={Ulm},
            postcode={89081}, 
            state={Baden-Württemberg},
            country={Germany}}

\begin{abstract}
Time series foundation models provide a universal solution for generating forecasts to support optimization problems in energy systems. Those foundation models are typically trained in a prediction-focused manner to maximize forecast quality.
In contrast, decision-focused learning directly improves the resulting value of the forecast in downstream optimization rather than merely maximizing forecasting quality. The practical integration of forecast values into forecasting models is challenging, particularly when addressing complex applications with diverse instances, such as buildings. This becomes even more complicated when instances possess specific characteristics that require instance-specific, tailored predictions to increase the forecast value. To tackle this challenge, we use decision-focused fine-tuning within time series foundation models to offer a scalable and efficient solution for decision-focused learning applied to the dispatchable feeder optimization problem. To obtain more robust predictions for scarce building data, we use Moirai as a state-of-the-art foundation model, which offers robust and generalized results with few-shot parameter-efficient fine-tuning. Comparing the decision-focused fine-tuned Moirai with a state-of-the-art classical prediction-focused fine-tuning Morai, we observe an improvement of 9.45\% in average total daily costs.
\end{abstract}

\begin{graphicalabstract}

    \centering
    \includegraphics[width=0.8\linewidth]{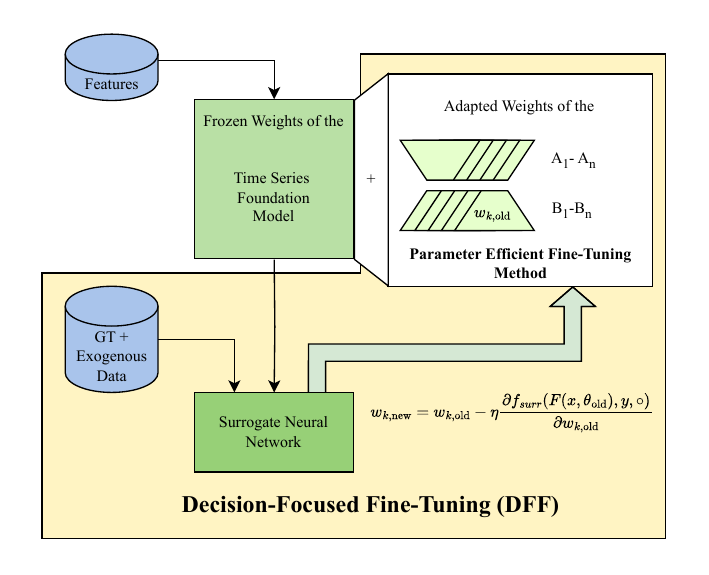}

\end{graphicalabstract}

\begin{highlights}

    \item This work presents the first approach that combines decision-focused learning with \acrfull{peft} methods to efficiently fine-tune pre-trained models.
    \item This is the first study to integrate decision-focused learning into fine-tuning time series foundation models to decrease downstream decision costs.
    \item This work is the first connecting a time series foundation model with an energy-related optimization problem and applying decision-focused learning.

\end{highlights}


\end{frontmatter}



\section{Introduction}

\begin{figure}
    \centering
    \includegraphics[width=0.7\linewidth]{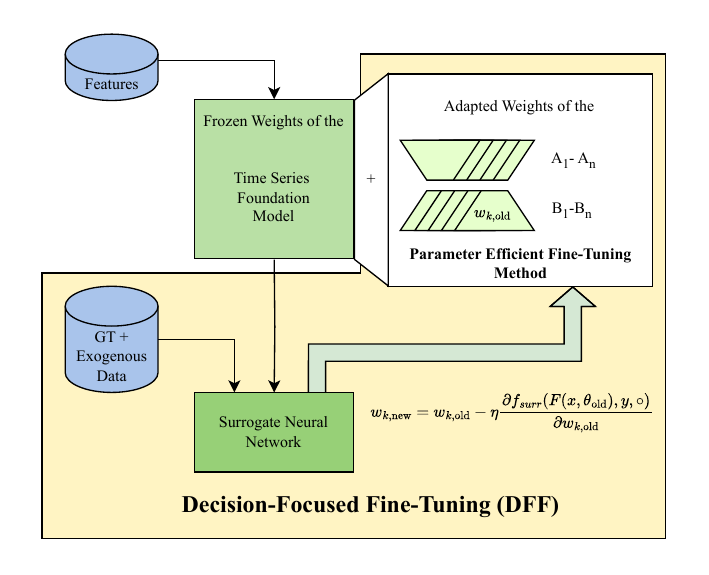}
    \caption{An overview of the decision-focused fine-tuning process which combines surrogate neural networks with \acrfull{peft} methods. It employs the technique from \cite{dfr} using a Surrogate Neural Network and its gradient to influence the weights of the \acrshort{peft} with respect to the downstream optimization.}
    \label{fig:graph_abstract}
\end{figure}

Many optimization applications from various domains, such as inventory optimization \cite{seyedan2023order}, portfolio optimization \cite{chan1999portfolio}, energy system optimization \cite{reddy2016day}, and building level schedule optimization \cite{appino_use_2018} rely on forecasts as input.
However, even though the forecasts are mainly generated as inputs for downstream optimizations, they are often evaluated and trained using forecast quality metrics, such as \acrfull{mse} or \acrfull{mae}. Those forecasts do not always align with the impact in subsequent or downstream optimization applications \cite{david_value_2021, Kourentzes_2020_inventory_planning, putz_true_2023, werling_impact_2023}. This impact can be denoted as the forecast value.

To address the challenge and divergence in objective, decision-focused learning aims to optimize the forecasting model to directly improve forecast value rather than merely maximizing forecast quality \cite{mandi_decision-focused_2024}. This approach focuses on forecast value and contrasts with the training paradigm of Prediction-Focused Learning. However, the practical integration of forecast value into learning models remains challenging, particularly when considering large-scale applications carried out on various instances, which need tailored forecasts related to the instance-specific properties.
An "instance" refers to a specific entity within a system that has unique properties that affect the optimization process. For example, in an energy system battery optimization problem such as the dispatchable feeder problem, each building is considered an instance with varying characteristics, such as its \acrlong{pv} installation and data, which influence the optimization outcomes \cite{werling_impact_2023}. \cite{dfr} proposes a scalable solution across instances of their optimization system by proposing a surrogate neural network within a Decision-Focused Retraining process to generate suitable forecasts. 

To optimize at the residual level, accurate and suitable forecasts at the household level are required in the energy sector. Recently, foundation models have gained traction in the load forecasting community \cite{stlf_foundation}, offering a robust solution to address data scarcity \cite{LIAO2025124973} while achieving state-of-the-art performance \cite{stlf_foundation}. These models serve as an excellent baseline for fine-tuning, having been trained prediction-focused across a diverse range of time series tasks. However, their out-of-the-box performance may not capture the unique instance-specific fluctuations that play an important role in household-level forecasts.

Therefore, this paper makes a threefold contribution at the intersection of decision-focused learning, foundation models, and energy forecasting. By integrating these domains, we address the need for decision-focused forecasting while leveraging the strengths of time series foundation models in an energy optimization use case context. The three contributions are:

First, this work introduces a novel approach that combines decision-focused learning with fine-tuning techniques applied to pre-trained models. Second, it is the first study to integrate decision-focused elements into the fine-tuning of time series foundation models. The resulting method, \acrfull{dff}, illustrated in \Cref{fig:graph_abstract}, builds on the retraining methodology proposed in \cite{dfr}, providing a scalable and efficient solution for energy systems with numerous instances. Third, within the energy community, this work is the first to connect time series foundation models with decision-focused learning within an energy-related optimization use-case. Furthermore, it demonstrates a global model that requires training only once, providing an efficient and scalable solution with improved performance compared to purely prediction-focused global models.

The remainder of this paper is structured as follows. First, a brief overview of related work is provided in \Cref{sec:rw}, covering decision-focused learning, foundation models, and fine-tuning methods. It is followed by a detailed description of the proposed decision-focused fine-tuning methodology in \Cref{sec:method}. An experiment applying \acrshort{dff} on an exemplary optimization problem and a study of data usage in different fine-tuning modes and two different fine-tuning methods are outlined in \Cref{sec:app} and \Cref{sec:imp}. The results are shown in \Cref{sec:res} and discussed in \Cref{sec:dis}. Finally, \Cref{sec:con} concludes the findings of this research.


\section{Related Work}
\label{sec:rw}

This section is fourfold. First, it introduces foundation models. Afterward, it introduces how those networks are adapted and finetuned using \acrfull{peft} approaches. Finally, it introduces the paradigm of learning to decision-focused learning.  It concludes by pointing out the contribution.

\subsection{Foundation Models}
Foundation models are used in various domains, such as computer vision \cite{awais2023foundational, ravi2024sam, xiao2024florence}, natural language processing \cite{myers2024foundation}, graph learning \cite{zhou2024comprehensive} and time series forecasting \cite{Liang_2024}. They are known for their broad generalizability. Currently, there are various foundation models for time series \cite{ansari2024chronoslearninglanguagetime, das2024decoderonlyfoundationmodeltimeseries, goswami2024momentfamilyopentimeseries, rasul2024lagllamafoundationmodelsprobabilistic, woo2024unified, garza2023timegpt}. Time-series foundation models are useful, especially as they can be used zero-shot without training. Information about such models' technical background and taxonomy is listed in \cite{Liang_2024}. A detailed comparison and benchmarking of different data usage in zero-shot and few-shot forecasting can be found in \cite{li2024foundtscomprehensiveunifiedbenchmarking}. In the energy domain, time series foundation models are evaluated on short-term load forecasting in \cite{ansari2024chronoslearninglanguagetime, meyer2024benchmarkingtimeseriesfoundation_stlf}. In short-term load forecasting, the foundation model Chronos \cite{ansari2024chronoslearninglanguagetime} achieves zero-shot performance comparable to specially trained State-of-the-Art models in short-term load forecasting \cite{stlf_foundation}. \cite{meyer2024benchmarkingtimeseriesfoundation_stlf} showed that time series foundation models can make good predictions in a short-term load forecasting case on a household level. Foundation models can be fine-tuned with available data to increase their task-specific performance \cite{han2024parameterefficientfinetuninglargemodels}.
The authors of \cite{LIAO2025124973} fine-tuned the foundation-model TimeGPT \cite{garza2023timegpt} for load-forecasting in a data-scarce environment.

\subsection{Parameter Efficient Fine-Tuning} 
With the rise of foundation models, methods for \acrfull{peft} are emerging. \acrshort{peft} methods are especially useful for efficiently fine-tuning large models, as they only adjust a few parameters, leading to efficient and cost-reduced training, as the zero-shot performance is improvable by adapting available data. Two reviews are given in \cite{han2024parameterefficientfinetuninglargemodels,xu2023parameterefficientfinetuningmethodspretrained}. In this paper, we focus on two exemplary reparameterized fine-tuning methods, adapting selected weights of the neural network architecture:

\paragraph{LoRA} (Low-Rank Adaptation) \cite{hu2021LoRA} uses two low-rank matrices, where $B\in \mathbb{R}^{d\times r}, A\in \mathbb{R}^{r\times k}$, and the rank $r \ll \min(d,k)$, with the original frozen weight matrix $W_0\in \mathbb{R}^{d\times k}$. The resulting matrix $W'$ can be derived through

\begin{equation}
    W'= W_0+\Delta W=W_0+BA.
\end{equation}

Therefore, weight updates in \acrshort{lora} result from adapting the trainable weights low-rank matrices $B$ and $A$.

\paragraph{DoRA}
\label{DoRA}
 (Directional Rank Adaptation) \cite{liu2024DoRA} builds on \acrshort{lora} by adapting the direction of pre-trained weight matrices. Similarly, as in \acrshort{lora}, weight updates are achieved using low-rank matrices $B$ and $A$. \acrshort{dora} extends this by introducing a trainable vector $m$, initialized with $\| W_0 \|_c \in \mathbb{R}^{1\times k}$ and $\| \cdot  \|_c$ being the vector-wise norm of a matrix across each column, scaling the direction of weight update. It keeps the pre-trained weight $W_0$ fixed and modifies by updating the weights in $BA$. 

 Within \acrshort{dora} the resulting weight matrices can be written as

\begin{equation}
\label{eq:DoRa}
W' = m \frac{W_0 + BA}{\| W_0 + BA \|_c}.    
\end{equation}

Therefore, weight updates in \acrshort{dora} result from adapting the trainable weights low-rank matrices $B$ and $A$ and the vector $m$.

\subsection{Decision-Focused Learning}
A review of decision-focused learning is given in \cite{mandi_decision-focused_2024}. The main goal of decision-focused learning is to optimize forecasts concerning the subsequent downstream optimization. \cite{dfr, bansal_taskmet_2023, chung_decision-aware_2022,  elmachtoub_smart_2020, lawless2022notetaskawarelossreweighing, shah_leaving_2023,  shah_decision-focused_2022, zhang_cost-oriented_2022, zharmagambetov_landscape_2023} are using the concept of surrogate losses to optimize the upstream forecaster. These surrogate losses form useful gradients concerning the optimization problems' decision costs and, therefore, can be used to learn the needs of downstream optimization. This can be accomplished by learning a parametrization of a convex loss function \cite{bansal_taskmet_2023, shah_leaving_2023, shah_decision-focused_2022}, by an optimal piecewise linear approximation method and the Huber norm embedding technique \cite{zhang_cost-oriented_2022}, by weighting the \acrshort{mse} \cite{chung_decision-aware_2022, lawless2022notetaskawarelossreweighing}, or by learning a non-parametrical surrogate landscape \cite{zharmagambetov_landscape_2023}. Decision-focused retraining \cite{dfr} learns a surrogate neural network and retrains the forecaster in a subsequent training process after the initial training.

\subsection{Own Contribution}
The contribution combines all the aspects given in this section by making a threefold contribution.
\begin{itemize}
    \item First, to the best of the authors' knowledge, this work presents the first approach that combines decision-focused learning with \acrshort{peft} to efficiently fine-tune pre-trained models.
    \item Second, this is the first study to integrate decision-focused learning into fine-tuning time series foundation models to decrease downstream decision costs.
    \item Third, within the energy community, this work is the first to connect a time series foundation model with an exemplary energy-related optimization problem, e.g. dispatchable feeder, and applying decision-focused learning.
\end{itemize}

\section{Decision-Focused Fine-Tuning}
\label{sec:method}
This section combines aspects of decision-focused learning using surrogate networks from \cite{dfr} and fine-tuning of foundation models using \acrshort{peft} methods. It introduces \acrfull{dff}, derived from the main objective of decision-focused learning. The \acrshort{dff} approach is applied on a pre-trained time series foundation model, similar to \cite{dfr} which uses pre-trained models.

Decision-focused learning aims to maximize the forecast value for downstream optimization. Specifically, we define the value function $V(F(x, \theta),y)$\footnote{The value function $V(F(x, \theta),y)$ returns the forecast value. Since, in our case, the value function is defined as system costs, minimizing the value function implies maximizing the forecast value for the downstream optimization.} that returns the forecast value with $F(x, \theta)$ being a forecaster $F$ of $y$, with features $x$, parameterized by $\theta$. 
Consequently, the primary objective of decision-focused learning is to find the best parametrization of $\theta$, i.e.,
\begin{equation}
\operatorname*{argmin}_\theta(V(F(x, \theta),y)).
\end{equation}
To minimize the function $V(F(x, \theta),y)$, decision-focused learning methods try to determine a gradient concerning the forecast value, 
\begin{equation}
\nabla V(F(x, \theta),y) = \frac{\partial V(F(x, \theta),y)}{\partial \theta}.
\end{equation}

The value function $V(F(x, \theta),y)$ is not differentiable as it needs the forecast applied within the optimization problem. In contrast, the value function can be modeled by a neural network, which we call a surrogate neural network and thus differentiable by design as in \cite{dfr}. This function can be used to replace $V(F(x, \theta),y)$ by $f_{surr}(F(x, \theta),y,\circ )$ with $\circ$ being exogenous factors influencing the outcome of the optimization problem.

Therefore, we can write the above-defined gradient as
\begin{equation}
\nabla V(F(x, \theta),y) = \frac{\partial V(F(x, \theta),y)}{\partial \theta} = \frac{\partial f_{surr}(F(x, \theta),y ,\circ )}{\partial \theta}.
\end{equation}

To adapt time series foundation models, we use a \acrshort{peft} method. Instead of adapting the original weight matrix of the foundation model, only the weights of the \acrshort{peft} method adapter are trained. Using backpropagation, the trainable weights $w_k$ of the \acrshort{peft} adapter are fitted to minimize the value function $V(F(x, \theta),y)$. I.e., using \acrshort{dora}, the weights of the low-rank matrices $B$ and $A$ and the vector $m$ (see \Cref{DoRA}) are updated.
The updated weights $w_{k,{\text{new}}}$ can be described as 

\begin{equation}
w_{k,{\text{new}}} = w_{k,{\text{old}}} - \eta \frac{\partial f_{surr}(F(x, \theta_{\text{old}}),y ,\circ )}{\partial w_{k, \text{old}}},
\end{equation}

with the learning rate $\eta > 0$. In the exemplary case of \acrshort{dora}, the weights in  $B$, $A$, and $m$ together with the not updated weights from the frozen weight matrix $W_0$ form the resulting parameters $\theta$ of the forecaster, see \Cref{eq:DoRa}. This leads to weights fitted to the forecast value and not to forecast quality. The surrogate neural network can also be viewed as a surrogate loss function. This serves as an analogy to a loss function of prediction-focused learning, but in contrast, this function is aligned with the decision cost in the downstream optimization.


\section{Application of Decision-Focused Fine-Tuning}
\label{sec:app}

This section investigates the degree to which \acrshort{dff} can increase the forecast value when using a forecast from a foundation model in a downstream optimization. \footnote{Code will be made available via GitHub upon acceptance.} To explore this, we focus on one exemplary optimization problem: the dispatchable feeder problem. This section mainly addresses the experimental setup, which is kept constant throughout. Therefore, this section first introduces the optimization problem briefly in \Cref{subsec:op}. Afterward, the resulting forecast value from the optimization problem and the data used to realize the optimization problem are presented in \Cref{sec:Used-Data}. We then describe the applied models and their training in \Cref{sec:Models}. In \Cref{sec:fc_sz} the forecasting scenario and the fine-tuning setting are outlined before the benchmark methods are listed in \Cref{sec:Benchmark}.

\subsection{Dispatchable Feeder Optimization Problem} 
\label{subsec:op}
The dispatchable feeder optimization problem \cite{appino_use_2018} is a hierarchical two-level non-convex optimization problem concerning the scheduling of a dispatchable feeder, i.e., a system consisting of residential power consumption, \acrshort{pv} and a battery able to adhere to an optimized schedule.
\cite{appino_use_2018, sossan_achieving_2016} define a dispatchable feeder as a system consisting of two main components. The first is the inflexible and variable energy prosumption of a residential building, defined as the residential load minus the power generated from the \acrshort{pv} system. The second is a flexible and controllable residential battery system, subject to its physical constraints. 
The dispatchable feeder leverages the battery system to effectively manage the grid interaction, addressing the uncertainties associated with consumption and generation. The daily operation is structured on two levels. A day-ahead dispatch schedule (DS) is created in the first level based on prosumption forecasts. In the second level, the actual dispatch is determined by considering the real prosumption minimizing any deviation from the DS. The optimization problem is the same as in \cite{dfr}.  A more detailed description and all mathematical symbols are listed in \Cref{sec:OP}. 

\paragraph{Forecast Value of the Dispatchable Feeder Optimization Problem}

\label{subsec:metrics}

We assess the forecast model's impact on the dispatchable feeder optimization problem using the daily total costs metric established in \cite{appino_use_2018}. Lower daily total costs indicate higher forecast value.

The total costs encompass both the DS costs $C_{\text{DS}}$ of the first-level optimization problem and the imbalance costs $C_{\text{Imb}}$ incurred due to deviations from the DS during actual dispatch, handled in the second-level optimization. The DS costs promote self-consumption and peak shaving and are modeled by the cost function
\begin{align}
\begin{split}
\label{eq: cost}
    C_{\text{DS}}\big(\tilde{P}_g^+(k), \tilde{P}_g^-(k)\big) &= c_{\text{q}}^+\cdot(\tilde{P}_g^+(k) \cdot\Delta t)^2 + c_{\text{l}}^+\cdot\tilde{P}_g^+(k) \cdot\Delta t  \\
    & + c_{\text{q}}^-\cdot(\tilde{P}_g^-(k)\cdot \Delta t )^2 + c_{\text{l}}^-\cdot\tilde{P}_g^-(k)\cdot\Delta t,
\end{split}
\end{align}
with the positive DS $\tilde{P}_{g}^+(k) \geq 0$, and the negative DS $\tilde{P}_{g}^-(k) \leq 0$, the cost coefficients $c_{\text{q}}^+$, $ c_{\text{l}}^+,$ $ c_{\text{q}}^-,$ $ c_{\text{l}}^-$ $ \in \mathbb{R}_{\geq 0}$ and the time interval $\Delta t$. The imbalance costs represent deviations between the scheduled and actual dispatch. With the deviation \(\Delta{P}_g(k)\) the imbalance costs can be expressed as
\begin{equation}
C_{\text{Imb}}\big(\Delta{P}_g(k)\big) = c_{\text{q}}^{\Delta} \cdot \left| \Delta{P}_g(k) \cdot \Delta t \right|^2 + c_{\text{l}}^{\Delta} \cdot \left| \Delta{P}_g(k) \cdot \Delta t \right|,
\end{equation}
with \(c_{\text{q}}^{\Delta} \in \mathbb{R}_{\geq 0}\) and \(c_{\text{l}}^{\Delta} \in \mathbb{R}_{\geq 0}\) being weighting coefficients.

The total costs are then calculated as:
\begin{equation}
\begin{split}
    C_{\text{T}}\big(\tilde{P}_g^+(k), \tilde{P}_g^-(k),\Delta{P}_g(k)\big) &= C_{\text{DS}}\big(\tilde{P}_g^+(k), \tilde{P}_g^-(k)\big) + \alpha \cdot C_{\text{Imb}}\big(\Delta{P}_g(k)\big),
\end{split}
\end{equation}
where \(\alpha \in \mathbb{R}_{\geq 0}\) is a scaling factor for the costs of the imbalance. We select $\alpha=10$ as in \cite{dfr, werling_impact_2023, remo_appino_storage_2018} to highly penalize the imbalance in this scenario. Daily total costs are computed by summing the total costs over each day. For evaluation, the costs are averaged over all days and buildings under consideration, which we call average total daily costs, as in \cite{dfr}.

\subsection{Used Data}
\label{sec:Used-Data}
The 'Ausgrid - Solar Home Electricity' dataset \cite{ratnam_residential_2017} is used for evaluation. It includes time series data on electricity consumption and solar generation from 300 residential buildings over three years, from July 1, 2010, to June 30, 2013. The data is recorded at 30-minute intervals using gross metering. The residential buildings, part of the Ausgrid electricity network and subject to the residential electricity tariff, are selected randomly.

We resample the data to an hourly resolution. The prosumption is calculated by subtracting the \acrshort{pv} power generation from the load. Unlike the approach in \cite{dfr}, we do not scale prosumption by a factor of 0.5. Furthermore, we divide the data into training, validation, and test sets using the first year (July 1, 2010, to June 30, 2011) as the training data set, the second as the validation data set (July 1, 2011, to June 30, 2012),  and the last year as the test data set (July 1, 2012, to June 30, 2013). Further, we split within the building dimension and use the first 50 buildings for fine-tuning different data usage strategies. In addition, the last 200 buildings were taken for method evaluation. The buildings 51-100 are omitted in evaluation as they are involved in surrogate network training.

\subsection{Used Models}
\label{sec:Models}
An overview of the models used is given in \Cref{tab:models}. As an exemplary foundation model, we use the Moirai foundation model \cite{woo2024unified}. As a replacement for the value function surrogate neural network, we use the same model architecture training configuration as in \cite{dfr}. In contrast to \cite{dfr}, we adapt the data to fit the data split of training and validation of the test buildings. The Moirai foundation model's large number of total parameters motivates using \acrshort{peft} methods, see \Cref{tab:models}.

\paragraph{Moirai Foundation Model}
Moirai is a universal forecasting model that demonstrates robust performance across diverse zero-shot scenarios \cite{woo2024unified}. Serving as an exemplary model within the foundation model class, we leverage the Moirai foundation model \cite{woo2024unified} in its base configuration to forecast hourly prosumption. Point forecasts are obtained from Moirai by taking the mean from the samples of the predictive distribution.

\paragraph{Surrogate Neural Network}

We use an ensemble of five surrogate neural networks, equivalent to \cite{dfr}. These surrogate networks are generated using the same data and data augmentation techniques described in \cite{dfr} on the first 50 buildings. The original data from the second 50 buildings are used for validation. We use the same time data split as described in \ref{sec:Used-Data}. An ensemble of five surrogate networks serves as proxy for $f_{surr}(F(x, \theta), y, \circ)$, using min-max scaled exogenous data (a uniformly drawn state of energy in surrogate training and a simulated state of energy (using \acrshort{mae} forecaster for schedule estimation and \acrshort{mse} forecaster for state of energy estimation) in fine-tuning, and mean, minimum, maximum, standard deviation of the noon hour). For a more detailed description of the training of the surrogate neural network, see \cite{dfr}, where the same setup is applied.

\begin{table}[h]
    \centering
    \caption{Used models with their respective number of parameters.}
    \begin{tabular}{@{}lll@{}}
        \toprule
        Model               & Parameters & References \\ \midrule
        Moirai Base                  & 91,357,728          & \cite{woo2024unified}       \\
        5 x Surrogate Neural Network     & 24,753              & \cite{dfr} \\ \bottomrule
    \end{tabular}
    \label{tab:models}
\end{table}

\subsection{Forecasting Scenario and Fine-Tuning Setting}
\label{sec:fc_sz}
An overview of the configurations used to train the \acrshort{peft} method is shown in \Cref{tab:training_settings}. Using Moirai, we forecast the upcoming 42 hours of prosumption, starting at noon. We take the last week of hourly historical prosumption as context without additional features. We take the mean to create a deterministic forecast out of the 100 samples generated by Moirai. Therefore, we configure Moirai using a context length of $168$, a forecast horizon of $42$, a patch size of $32$, and a sample size of $100$. For the \acrshort{peft} adapter, we use an AdamW \cite{loshchilov2019decoupledweightdecayregularization} as an optimizer with a learning rate of 0.0001. We use a batch size of 32 with bfloat16 mixed-precision \cite{narang2017mixed} over 5 epochs.

\begin{table}[ht]
    \centering
    \caption{Moirai configuration and basic fine-tuning settings.}
    \begin{tabular}{@{}lll@{}}
        \toprule
        & Setting                    & Value \\ 
        \midrule
        \multirow{4}{*}{Moirai} & Patch Size & 32 \\ 
        & Context Length & 168 \\ 
        & Prediction Length & 42 \\ 
        & Numbers of Samples & 100 \\ 
        \midrule
        \multirow{2}{*}{Optimizer} & Optimizer                   & AdamW \\
        & Learning Rate               & 0.0001 \\

        \midrule
        \multirow{3}{*}{Fine-tuning} & Batch Size                  & 32 \\
        & Number of Epochs            & 5 \\ 
        & Precision & bfloat16 \\
        \bottomrule
    \end{tabular}
    \label{tab:training_settings}
\end{table}

\subsection{Benchmark Methods}
\label{sec:Benchmark}

To evaluate and compare the result of \acrshort{dff}, we use several benchmarks, as described below.

\paragraph{Prediction-Focused Fine-Tuning (PFF)} These models align more closely with traditional forecasting methods. Their use to optimize for downstream optimizations suggests that better forecast quality, measured by metrics such as \acrshort{mae} and \acrshort{mse}, is correlated with improved forecast values.

The \acrshort{mae} as loss function refers to the absolute error and minimizes the residuals. It is not sensitive to outliers. Mathematically, it can be written as
\[
\text{MAE} = \frac{1}{n} \sum_{i=1}^{n} | y_i - \hat{y}_i |
\]
referring to $y_i$, as the ground truth, $\hat{y}_i$ as the forecast and $n$ as the number of samples.

The \acrshort{mse} metric is more sensitive to outliers, as it weights the errors quadratically. It can be written as 
\[
\text{MSE} = \frac{1}{n} \sum_{i=1}^{n} (y_i - \hat{y}_i)^2
\]
with also referring to $y_i$, as the ground truth, $\hat{y}_i$ as the forecast and $n$ as the number of samples.

\paragraph{Zero-Shot (ZS)} The direct application of the Moirai model without fine-tuning. This represents a traditional zero-shot forecasting approach in the foundation model community.

\paragraph{Naive 48 (N48)} We use the last fully observed matching period as a forecast. Thus, the forecasts for the current horizon are based on the values observed 48 hours earlier to fit in a full forecast horizon of 42 and adjust it to daily patterns with forecasting at noon. This method assumes that the conditions 48 hours ago are similar to the current period.

\paragraph{Naive 168 (N168)} This method follows the same idea as Naive 48, but it considers weekly periodicity. The forecast is based on values from the same time exactly one week earlier, assuming that last week's observations closely match the current period.

\paragraph{Perfect Forecast (P-FC)} This method uses the actual observed values as the forecast. While this leads to a perfect forecast with the lowest possible cost, it is an unrealistic assumption because energy load and \acrshort{pv} generation cannot be predicted without errors due to their uncertain nature. Since there is no deviation from the schedule, no imbalance costs arise, as they only occur due to mismatches between the forecast and actual values.


\section{Impact of Fine-Tuning Modes and PEFT Methods in Decision-Focused Fine-Tuning}
\label{sec:imp}
 
This section introduces the varied settings of the fine-tuning modes and the \acrshort{peft} methods. Overall, the settings are systematically varied to assess their impact within the \acrshort{dff} method. The first setting relates to the data usage in fine-tuning, and the second explores two versions of \acrshort{peft} methods, which are introduced in \Cref{sec:rw}. The varied settings are also performed using the benchmark methods, if possible, to give a complete overview.

\subsection{Comparison of Different Fine-Tuning Modes}

\begin{figure}
    \centering
    \includegraphics[width=1\linewidth]{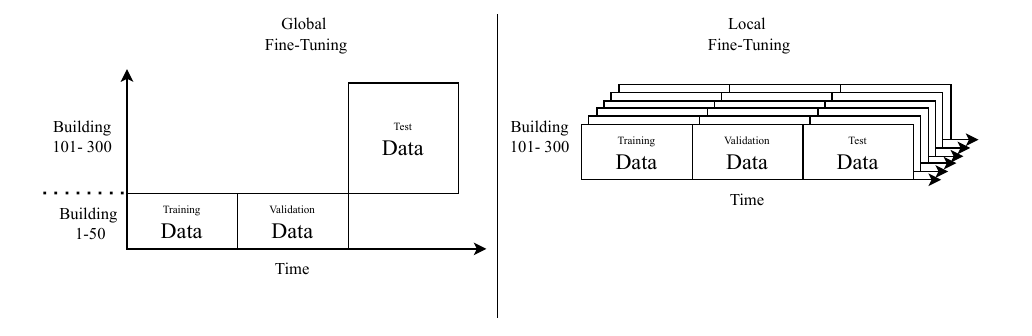}
    \caption{Visualization of the global and local fine-tuning mode. The global mode performs training by considering a set of buildings (1-50), whereas the local mode performs training by considering the buildings 101-300 independently. For both methods, the test sets are identical. The buildings 51-100 are omitted in evaluation as they are involved in surrogate network training. The buildings 51-100 are omitted in evaluation as they are involved in surrogate network training.}
    
    \label{fig:fine-tuning_mode}
\end{figure}

We compare two fine-tuning paradigms: global fine-tuning and local fine-tuning. Those paradigms focus mainly on the use of data within model fine-tuning. In our case, the building instances used in fine-tuning differ. These concepts are visualized in Figure \ref{fig:fine-tuning_mode}. 

\paragraph{Global Fine-Tuning}

Global fine-tuning refers to fine-tuning a model using data that are not directly related to the specific instance being forecasted. This approach is closely related to zero-shot forecasting, where no customization or fine-tuning is performed for the particular test instance and data. In our case, the model is fine-tuned on data from the first 50 buildings. Strong global model performance suggests the model can generalize well for the optimization task across different instances, even without instance-specific fine-tuning.

\paragraph{Local Fine-Tuning}

In contrast to global fine-tuning, local fine-tuning uses data specific to the forecasted instance, such as an individual building. This approach leverages instance-specific data, aligning the forecasting model with that instance's unique characteristics and patterns and its optimization context. Compared to global learning, this tailored method is expected to enhance performance and suitability instance-specific for downstream optimization, as it directly optimizes for the instance in question.

\subsection{Comparison of PEFT Methods}

The two different configurations used for the \acrshort{peft} adapter can be seen in \Cref{tab:LoRA_DoRA_comparison}. To compare those two techniques, we keep the basic parameter of \acrshort{lora} equal to \acrshort{dora}. Therefore, $r = 8$ is chosen, and $\alpha = 32$. We apply no dropout or additional regularization techniques and only adjust Moirai Network's transformer blocks ($v\_proj$,  $q\_proj$, $k\_proj$, and $out\_proj$). This results in \acrshort{dora} having 626,688 parameters compared to \acrshort{lora} with 589,824 parameters. Independently of the chosen \acrshort{peft} method, we only adjust a small number of parameters, i.e., less than 0.7\% of the total network parameters.

\begin{table}
    \centering
    \caption{Comparison between \acrshort{lora} and \acrshort{dora}. In the top part, the configuration is given in the lower part the number of parameters of the resulting network is shown.}
    \begin{tabular}{@{}lll@{}}
        \toprule
        Aspect             & \acrshort{lora}               & \acrshort{dora}              \\\midrule

        \multicolumn{1}{l}{\acrshort{lora} Configuration} & $r = 8$                  & $r = 8$                   \\ 
                                    & $\alpha = 32$        & $\alpha = 32$       \\ 
                                    & $Dropout = 0.0$    & $Dropout = 0.0$    \\ 
        
        \multicolumn{1}{l}{Target Modules}                & $v\_proj$, $q\_proj$,  & $v\_proj$, $q\_proj$, \\
        & $k\_proj$, $out\_proj$ & $k\_proj$, $out\_proj$ \\
        \midrule
        Trainable Parameters          & 589,824                    & 626,688                   \\ 
        Total Parameters              & 91,947,552                 & 91,984,416                \\ 
        Trainable Percentage          & 0.6415\%                  & 0.6813\%                 \\ 
        \bottomrule
    \end{tabular}
    \label{tab:LoRA_DoRA_comparison}
\end{table}


\section{Results}
\label{sec:res}

The presented results are the mean and standard deviation across five independent runs concerning the 200 test buildings. \Cref{tab:mean_df_run} provides a detailed comparison of various model configurations using three key performance indicators: Average Daily Total Costs (€) as the optimization forecast value, \acrshort{mae} in \si{(\kilo\watt)}, and \acrshort{mse} in \si{(\kilo\watt)\squared} as two forecast quality metrics. 

\begin{table}[p]
\centering
\caption{Configuration Specific Performance Evaluation with Mean and Standard Deviation. Sorted by the Performance within the Forecast Value Average Daily Total Cost. }

\label{tab:mean_df_run}
\begin{tabular}{llllcccccc}
\toprule
& \multicolumn{3}{c} {Configuration} & \multicolumn{6}{c}{Performance Factor}  \\
Name & FT. & Loss & \acrshort{peft} & \multicolumn{2}{c}{\makecell{A. D. Total \\ Costs (\euro{})}} & \multicolumn{2}{c}{\acrshort{mae} (\si{\kilo\watt})} & \multicolumn{2}{c}{\acrshort{mse} (\si{(\kilo\watt)\squared})} \\
     & Mode & Func. & Method & Mean  & Std   & Mean  & Std   & Mean  & Std \\
\midrule
P-FC &   &       &   & \textbf{4.18}  &     &     &     &     &  \\
\midrule
\midrule
\acrshort{dff} & Local & Surr. & \acrshort{dora}  & \textbf{12.75} & 0.04  & 0.46   & 0.00  & 0.53   & 0.00 \\
\acrshort{dff} & Local & Surr. & \acrshort{lora}  & 12.79  & 0.03   & 0.46 & 0.00  & 0.53 & 0.00 \\
\acrshort{dff} & Global& Surr. & \acrshort{lora}  & 13.15  & 0.14   & 0.57   & 0.01   & 0.73   & 0.03 \\
\acrshort{dff} & Global& Surr. & \acrshort{dora}  & 13.18  & 0.20   & 0.57   & 0.01   & 0.73   & 0.03 \\
 \acrshort{pff} & Local & \acrshort{mse}       & \acrshort{lora}  & 14.08  & 0.05   & 0.38   & 0.00  & \textbf{0.41}   & 0.00 \\
 \acrshort{pff} & Local & \acrshort{mse}       & \acrshort{dora}  & 14.09  & 0.02   & 0.38   & 0.00  & \textbf{0.41}   & 0.00 \\
 \acrshort{pff} & Global& \acrshort{mse}       & \acrshort{dora}  & 14.13  & 0.37   & 0.38   & 0.01   & \textbf{0.41}  & 0.00 \\
 \acrshort{pff} & Global& \acrshort{mse}       & \acrshort{lora}  & 14.26  & 0.26   & 0.38   & 0.00  & \textbf{0.41}   & 0.00 \\
\midrule
ZS  &  &       &   & 15.99  &     & 0.42   &     & 0.50   &  \\
N48     &   &       &   & 16.77  &     & 0.45   &     & 0.69   &  \\
\midrule
\acrshort{pff} & Local & \acrshort{mae}       & \acrshort{lora}  & 17.53  & 0.04  & \textbf{0.36}   & 0.00  & 0.43   & 0.00 \\
 \acrshort{pff} & Local & \acrshort{mae}       & \acrshort{dora}  & 17.57  & 0.05   & \textbf{0.36}  & 0.00  & 0.43   & 0.00 \\
\midrule
N168    &   &       &   & 17.83  &     & 0.45   &     & 0.69   &  \\
\midrule
\acrshort{pff} & Global& \acrshort{mae}       & \acrshort{lora}  & 18.45  & 0.26   & 0.36   & 0.00  & 0.43   & 0.00 \\
\acrshort{pff} & Global& \acrshort{mae}       & \acrshort{dora}  & 20.37  & 4.35   & 0.37   & 0.03   & 0.47   & 0.07 \\
\bottomrule
\multicolumn{10}{p{\textwidth}}{\scriptsize 
    Abbreviations: \acrshort{dff}: \acrlong{dff}; \acrshort{pff}: \acrlong{pff}; P-FC: Perfect Forecast; ZS: Zero-Shot; N48: Naive 48; Naive N168; A. D. Total Costs: Average Daily Total Costs; 
    \acrshort{peft}: \acrlong{peft}; FT: Fine-Tuning; Surr: Surrogate Network; 
    \acrshort{mae}: \acrlong{mae}; \acrshort{mse}: \acrlong{mse}; 
    \acrshort{dora}: \acrlong{dora}; \acrshort{lora}: \acrlong{lora}; 
    Std: Standard Deviation; Func: Function.
}
\end{tabular}
\end{table}
\begin{figure}
    \centering
    \includegraphics[width=1\linewidth]{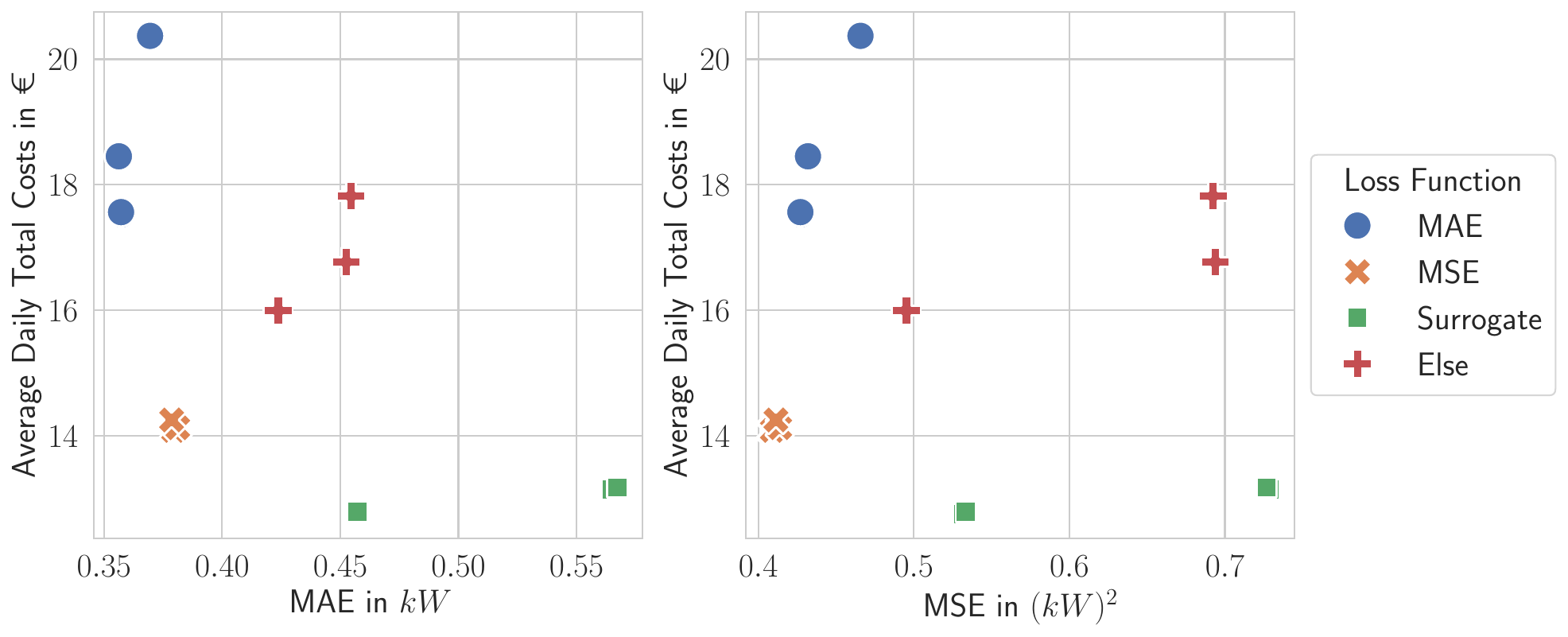}
    \caption{Average Daily Total Costs created by the different loss functions used versus the forecast quality metrics \acrshort{mse} and \acrshort{mae}. Else refers to the Benchmarks, but the Perfect Forecast Benchmark is excluded.}
    \label{fig:scatter_loss_function_MAE_mse_vs_costs}
\end{figure}

A visualization showing the relation of forecast value and the forecast quality metrics \acrshort{mae} and \acrshort{mse} are shown in \Cref{fig:scatter_loss_function_MAE_mse_vs_costs}. It especially visualizes the relation between the loss and the resulting total costs. It shows further that the choice of the loss function leads to the respective minimization of the performance factor. This visualization is additionally available for different fine-tuning modes and \acrshort{peft} methods, see \Cref{fig:forecast_value_vs_forecast_quality} in the \Cref{sec:Appendix_plots}.

\paragraph{Forecast Value} 
The Perfect Forecast benchmark demonstrates the lowest average daily total cost, representing an ideal scenario with no discrepancies between forecasted and actual values. With average daily costs as low as €4.18, this benchmark allows the system to consistently operate at this minimal cost. While it serves as a valuable point of comparison, it is essential to note that achieving such perfection is not possible in real-world applications. Regarding Average Daily Total Costs, the \acrshort{dff} configurations trained locally using the surrogate neural network show costs of €12.75 for \acrshort{dora} and €12.79 for \acrshort{lora}. In contrast, the costs for the global fine-tuning mode are €13.18 for \acrshort{dora} and €13.15 for \acrshort{lora}. The \acrshort{pff} models fine-tuned with the \acrshort{mse} loss function leads to costs of €14.09 for local modes and up to €14.26 for global modes. Models that utilize the \acrshort{mae} loss function have higher costs; the best local model reaches €17.53, while the global \acrshort{dora} model reaches €20.37. Notably, \acrshort{pff} fine-tuning using the \acrshort{mae} loss function results in increased total costs compared to the zero-shot usage of Moirai and the Naive 48 benchmark. Furthermore, \acrshort{pff} models trained globally with the \acrshort{mae} loss function perform worse than the Naive 168 benchmark. Comparing the best \acrshort{pff} method (€14.08) with the best \acrshort{dff} (€12.75) method, we observe an improvement of 9.45\% in average total daily costs. If we compare it against the non-fine-tuned foundation model (€15.99) we improve over 20.26\% in average total daily costs.

\paragraph{Forecast Quality} For \acrshort{mae}, models fine-tuned with the \acrshort{mae} loss function show values of \SI{0.36}{\kilo\watt} for both \acrshort{lora} and \acrshort{dora} across local and global fine-tuning modes. Models fine-tuned with the surrogate neural network show \acrshort{mae} values around \SI{0.46}{\kilo\watt} for local fine-tuning modes and \SI{0.57}{\kilo\watt} for global fine-tuning modes. When evaluating \acrshort{mse}, models fine-tuned with the \acrshort{mse} loss function show values around \SI{0.41}{(\kilo\watt)\squared}, consistent across both local and global modes. Models fine-tuned with the surrogate neural network show \acrshort{mse} values around \SI{0.53}{(\kilo\watt)\squared} for local fine-tuning modes and 0.73 \si{(\kilo\watt)\squared} for global modes.


\section{Discussion}
\label{sec:dis}
In this section, we discuss four different topics. First, we want to discuss the usage of \acrshort{lora} compared to \acrshort{dora}. Afterward, we elaborate on the choice of the loss function, discuss global and local fine-tuning, and conclude on the potential of \acrshort{dff}. We finish this section by pointing out future work.

\paragraph{PEFT Methods} \acrshort{dora} and \acrshort{lora} show comparable performance in most metrics, with only minor differences. Both methods effectively fine-tune and work across all metrics and their respective loss functions. They are both effective within the \acrshort{dff} methodology. The current study does not provide definitive guidance on choosing \acrshort{dora} over \acrshort{lora}. However, regarding parameter efficiency, \acrshort{lora} demonstrates greater effectiveness, because a smaller amount of trainable weights is used \Cref{tab:LoRA_DoRA_comparison}.

\paragraph{Loss Functions} The choice of loss function affects the model's performance. This is especially visible in \Cref{fig:scatter_loss_function_MAE_mse_vs_costs}, where the non-linear dependency of the forecast quality and the forecast value is shown. The surrogate neural networks, aligned with the target metric of minimizing average daily total costs, yield the best results in terms of cost reduction across both fine-tuning modes but also lead to a deterioration in terms of \acrshort{mse} and \acrshort{mae}. \acrshort{mse} and \acrshort{mae}, while useful for minimizing their specific error metrics, lead to higher total costs. \acrshort{mse} is better at managing large deviations, while \acrshort{mae} lowers overall error but incurs the highest costs, especially in global fine-tuning mode. This may result from high imbalance costs, which arise due to outliers within the \acrshort{mae} forecast, which cannot be compensated by flexibility in the battery anymore. Outliers are less prevalent in the \acrshort{mse} forecast, making it more suitable for this optimization problem.

\paragraph{Global vs. Local Fine-Tuning} Local fine-tuning emphasizes the individual building, improving cost efficiency. However, global fine-tuning offers scalability and generalization across unseen instances, such as buildings, making it more suitable for real-world applications. Although global fine-tuning tends to have higher costs, its ability to generalize across unseen buildings without requiring extensive fine-tuning makes it a more scalable solution for broader deployment, as the training only needs to be done once and does not scale with the number of instances. Thus, depending on the savings and the training costs, local or global fine-tuning might be beneficial.

\paragraph{Potential of DFF} We observe that fine-tuning within this domain can lead to better decision costs, as we observe an improvement of 20.26\% compared to the not fine-tuned model within the zero-shot task. In comparison to \acrshort{pff} models, the improvement also seems promising, with an increase of 9.45\%.

\paragraph{Future Work}

Despite the study showing positive results regarding the mean performance across 200 buildings, the averaging process does not examine the impact on individual buildings in detail. Therefore, this approach may be better suited for buildings well-modeled by the surrogate. Since the behavior of each building is not examined in detail, further insights into individual building performance could potentially improve the method.

Additionally, as the approach in \cite{dfr}, this method is not tailored to accommodate a specific convex loss function, meaning the surrogate function does not ensure convexity, which may result in unstable solutions across different buildings. Integrating additional building variables as exogenous features could improve stability by capturing unique dynamics for each building. Moreover, adapting the model to parameterize a convex loss function, as proposed in \cite{bansal_taskmet_2023, shah_leaving_2023,shah_decision-focused_2022}, might further enhance stability during the fine-tuning process.

Within the \acrshort{dff} method, a comparison of different variants of the \acrshort{peft} methods and foundation networks or the application of different surrogate neural networks or losses with different properties could be investigated. Further, the current forecasting method does not adhere to exogenous features, this could lead to increased model performance and stabilize the forecaster itself. Further, the potential of global fine-tuning should be evaluated in detail, as it offers a scalable solution across various instances.


\section{Conclusion}
\label{sec:con} 
One challenge with integrating optimization problems within the energy systems is the generation of suitable forecasts. This paper begins by identifying three domains: decision-focused learning to generate suitable forecasts, \acrfull{peft} to efficiently fine-tune models, and time series foundation models as a baseline for fine-tuning. It demonstrates the potential of combining these domains by adapting the Decision-Focused Retraining methodology of \cite{dfr} for fine-tuning of time series foundation models using \acrshort{peft}. This paper evaluates the \acrfull{dff} methodology based on the dispatchable feeder optimization problem, which is a typical battery scheduling optimization problem. The results show that both the \acrfull{dora} and \acrfull{lora}  methods are effective with minimal differences in performance. \acrshort{dff} notably enhances model performance, with the surrogate neural network demonstrating a clear advantage in minimizing decision costs. Specifically, \acrshort{dff} outperforms \acrfull{pff} by more than 9.45\% in the selected exemplary dispatchable feeder problem. Furthermore, while global \acrshort{dff} leads to more effortless scalability and generalization, local instance-specific \acrshort{dff} remains more cost-effective. Future work should focus on developing scalable strategies to refine the global data strategy, making decision-focused learning methods more applicable to real-world scenarios.

\section{Declaration}
During the preparation of this work the authors used ChatGPT in order to improve clarity and fluency during the writing process. After using this tool/service, the authors reviewed and edited the content as needed and take full responsibility for the content of the published article.

\section{Acknowledgments}
This project is funded by the Helmholtz Association’s Initiative and Networking Fund through Helmholtz AI, the Helmholtz Association under the Program “Energy System Design”, and the German Research Foundation (DFG) as part of the Research Training Group 2153 “Energy Status Data: Informatics Methods for its Collection, Analysis and Exploitation”. This work is supported by the Helmholtz Association Initiative and Networking Fund on the HAICORE@KIT partition.


\clearpage
\appendix
\section{The Dispatchable Feeder Optimization Problem}
\subsection{Problem Formulation}
\label{sec:OP} The mathematical formulation and the parameters are the same as in \cite{dfr}.
The entire operation can be defined as a hierarchical two-level non-convex optimization problem. Within both levels, the time operation is discretized into time intervals indexed by $k \in \mathbb{N}$ with an interval length of $\Delta t \in \mathbb{R}$.
The grid interaction is represented solely by the active power exchange with the dispatchable feeder, comprising the first component, the battery power input and the second component, the uncertain prosumption from the building, summed together. The flexible battery is modeled by its active power input $P_s(k) \in [\underline{P_s}, \overline{P_s}]$ and state of energy $E_s(k) \in [\underline{E_s}, \overline{E_s}]$ with lower and upper bounds $\underline{P_s}, \overline{P_s} \in \mathbb{R}$ and $\underline{E_s}, \overline{E_s} \in \mathbb{R}_{\geq 0}$. The progression of the state of energy of the battery is formulated as
\begin{align}
\label{eq: modified SOC}
    E_s(k+1) = E_s(k) + \Delta t \cdot \big(P_s(k) - \mu P^+_s(k) + \mu P^-_s(k)\big)
\end{align}
with loss coefficient $0 \leq \mu \leq 1$ and positive and negative directions of the battery's power input $P^+_s(k) \geq 0$ and $P^-_s(k) \leq 0$.

Based on this description of the two components, the optimization problem's two levels can be expressed as follows.

\paragraph{First Level: Computation of Dispatch Schedule}
The Dispatch Schedule (DS), is calculated within the first level. The DS $\tilde{P}_{g}(k) \in \mathbb{R}$ is calculated regarding the costs for the following day and considers point forecast of the prosumption $\hat{P}_l(k) \in \mathbb{R}$. 
The first-level optimization problem is then formulated as
\begin{align}
\begin{split}
\label{eq: opti}
 \!\min_{\{X\}_{\mathcal{K}}} \sum_{k \in \mathcal{K}} & C_{DS}\big(\tilde{P}_g^+(k), \tilde{P}_g^-(k)\big) \\ 
\text{s.t.} \ \text{for} & \text{ all}\ k \in \mathcal{K} \\
 & (\ref{eq: modified SOC}) \\ 
 E_s(k_0) &= E_s^0 \\
\tilde{P}_g(k) &= P_s(k) + \hat{P}_l(k) \\
  \tilde{P}_g(k) &= \tilde{P}_g^+(k) + \tilde{P}_g^-(k) \\
 \tilde{P}_g^+(k) &\geq 0 \\
 \tilde{P}_g^-(k) &\leq 0 \\
  P_s(k) &= P_s^+(k) + P_s^-(k) \\
 P_s^+(k) &\geq 0\\
 P_s^-(k) &\leq 0\\
 0 &= P_s^+(k) \cdot P_s^-(k) \\
 \underline{P_s} \leq &P_s(k) \leq \overline{P_s} \\
\underline{E_s} \leq &E_s(k+1) \leq \overline{E_s} 
\end{split}
\end{align} 
with a discrete scheduling horizon $\mathcal{K}$, decision vector $X(k) = \big(\tilde{P}_g(k),$ $\tilde{P}_g^+(k),$ $\tilde{P}_g^-(k), E_s(k+1), P_s(k),$ $P_s^+(k), P_s^-(k)\big)^T$, parameters $E_s^0,$ $\underline{P_s}, \overline{P_s},$ $\underline{E_s}, \overline{E_s}$, and point forecasts $\hat{P}_l(k)$. Note, it is necessary to know or estimate the state of energy at the beginning of scheduling $k_0 \in \mathbb{N}$.

\paragraph{Second Level: Calculation of the Actual Dispatch}
After the computation of the DS, in the second level, the actual dispatch is calculated for every time interval based on the actual prosumption $P_l(k) \in \mathbb{R}$. Within this level, the aim is to minimize the deviation of the corresponding computed DS $\Delta P_g(k) \in \mathbb{R}$ within the technical constraints. In our problem, the second-level optimization problem can be formulated as 
\begin{align}
\begin{split}
\label{eq: online}
 \!\min_{X(k)} \ & \big(\Delta P_g(k)\big)^2 \\
 & (\ref{eq: modified SOC}) \\ 
 E_s(k) &= E_s^k \\
P_g(k) &= P_s(k) + P_l(k) \\
  P_g(k) &= \tilde{P}_g(k) + \Delta {P}_g(k) \\
  P_s(k) &= P_s^+(k) + P_s^-(k) \\
 P_s^+(k) &\geq 0\\
 P_s^-(k) &\leq 0\\
 0 &= P_s^+(k) \cdot P_s^-(k) \\
 \underline{P_s} \leq &P_s(k) \leq \overline{P_s} \\
\underline{E_s} \leq &E_s(k+1) \leq \overline{E_s} 
\end{split}
\end{align} 
with decision vector $X(k) = \big(P_g(k),$ $E_s(k+1), P_s(k),$ $P_s^+(k), P_s^-(k)\big)^T$, parameters $\Tilde{P}_g(k), $ $P_l(k),$ $E_s^k,$ $\underline{P_s}, \overline{P_s},$ $\underline{E_s}, \overline{E_s}$, and actual dispatch $P_g(k) \in \mathbb{R}$. Note, the state of energy in $k \in \mathbb{N}$ is known.

\subsection{Parameter Specifications of the Dispatchable Feeder Optimization Problem}
\begin{table} 
\centering
    \caption{Chosen parameter specifications concerning the given optimization problem}
        \begin{tabular}{cc}
        \toprule
        Parameter & Value \\ 
        \midrule
        $\Delta t$ & \SI{1}{\hour}\\
        $\mathcal{K}$ & $\{k_s,...,k_s+29\}$  $^1$ \\
        $c_{q}^+$ & \SI{0.05}{(\EUR\per(\kilo\watt\hour)\squared)}\\
        $c_{l}^+$ & \SI{0.3}{(\EUR\per(\kilo\watt\hour))}  \\
        $c_{q}^-$ & \SI{0.05}{(\EUR\per(\kilo\watt\hour)\squared)}\\
        $c_{l}^-$ & \SI{0.15}{(\EUR\per(\kilo\watt\hour))}  \\
        $\underline{P_{s}}$ & \SI{-5}{(\kilo\watt)}\\
        $\overline{P_{s}}$ & \SI{5}{(\kilo\watt)}\\
        $\underline{E_{s}}$ & \SI{0}{(\kilo\watt\hour)}\\
        $\overline{E_{s}}$ & \SI{13.5}{(\kilo\watt\hour)}\\
        $\mu$ & $0.05$ \\
        $E_s^0$ & day 1: \SI{6.75}{(\kilo\watt\hour)}\\
        $\alpha$ & $10$ {\cite{remo_appino_storage_2018}} \\
        $c_{\text{q}}^{\Delta} $& \SI{0.05}{(\EUR\per(\kilo\watt\hour)\squared)}\\
        $c_{\text{l}}^{\Delta} $ &\SI{0.3}{(\EUR\per(\kilo\watt\hour))} \\
        \bottomrule
        \multicolumn{2}{l}{\scriptsize $^1$ $k_s \in \mathbb{N}$ starts at midnight and the schedule calculation at noon}\\
        \end{tabular}

    \label{tab: opti parameter}
\end{table}

\Cref{tab: opti parameter} shows the parameter specifications of the dispatchable feeder optimization problem. Note that the parameter $\alpha$ is set to 10 as in \cite{dfr, remo_appino_storage_2018}, which results in a high weighting of the imbalance costs and, thus, high costs for deviations from the DS. Therefore, the main aim in terms of cost reduction is to reduce the occurrence of imbalances.

\clearpage
\section{Supplementary Result Plots}
\label{sec:Appendix_plots} 
\begin{figure}[h]
    \centering
    \includegraphics[width=0.73\linewidth]{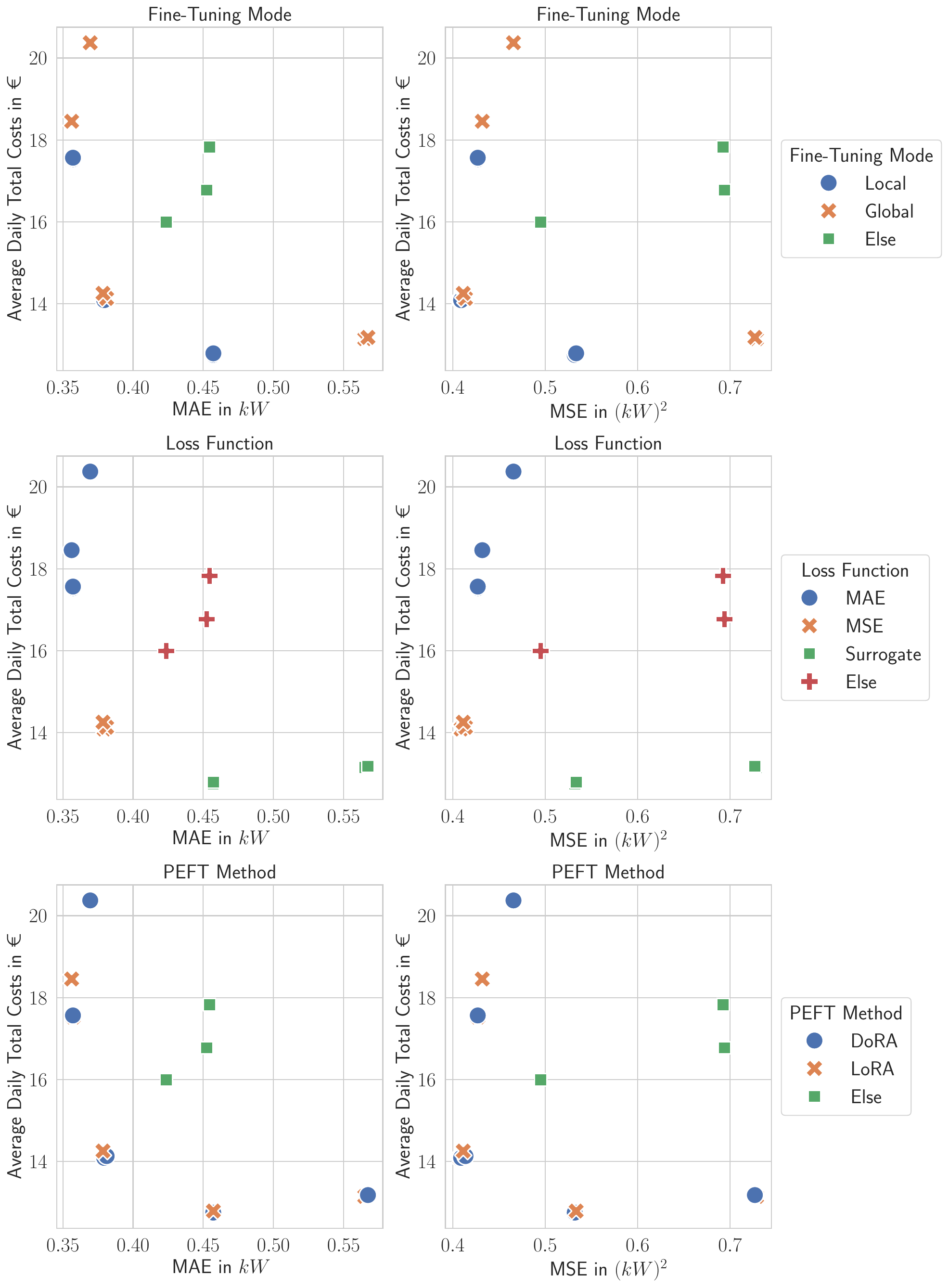}
    \caption{A comparison of forecast value and quality concerning the three factors (Loss Function, Fine-Tuning Mode and PEFT Method).  
   }
    \label{fig:forecast_value_vs_forecast_quality}
\end{figure}


\end{document}